\documentclass[letterpaper, 10 pt, conference]{ieeeconf}  

\IEEEoverridecommandlockouts                              

\usepackage{graphics} 
\usepackage{subfigure}
\usepackage{epsfig} 
\usepackage{mathptmx} 
\usepackage{times} 
\usepackage{amsmath} 
\usepackage{amssymb}  
\usepackage{booktabs}
\usepackage{makecell}
\usepackage{hyperref}
\usepackage{bbding}
\usepackage{color}
\usepackage[symbol]{footmisc}
\title{\LARGE \bf
Omnidirectional Depth-Aided Occupancy Prediction based on Cylindrical Voxel for Autonomous Driving
}

\author{Chaofan Wu$^{\ast1}$ , Jiaheng Li$^{\ast1}$, Jinghao Cao$^{1}$, Ming Li$^{1}$, Yongkang Feng$^{1}$, Jiayu Wu$^{1}$\\Shuwen Xu$^{1}$, Zihang Gao$^{1}$, Sidan Du\textsuperscript{\Envelope}$^{1}$, Yang Li\textsuperscript{\Envelope}$^{1,2}$
\thanks{*: Equal Contribution}
\thanks{\Envelope: Corresponding Author}}

\begin{document}
\bibliographystyle{unsrt}

\maketitle
\thispagestyle{empty}
\pagestyle{empty}

\begin{abstract}
Accurate 3D perception is essential for autonomous driving. Traditional methods often struggle with geometric ambiguity due to a lack of geometric prior. To address these challenges, we use omnidirectional depth estimation to introduce geometric prior. Based on the depth information, we propose a Sketch-Coloring framework OmniDepth-Occ. Additionally, our approach introduces a cylindrical voxel representation based on polar coordinate to better align with the radial nature of panoramic camera views. To address the lack of fisheye camera dataset in autonomous driving tasks, we also build a virtual scene dataset with six fisheye cameras, and the data volume has reached twice that of SemanticKITTI. Experimental results demonstrate that our Sketch-Coloring network significantly enhances 3D perception performance. 
\end{abstract}

\section{INTRODUCTION}
Accurate perception of the surrounding environment is crucial for autonomous driving system. In recent years, vision-based 3D perception algorithms have gained significant attention and advancement. The typical workflow involves employing a 2D encoder to extract latent representations from images and a view transformation method~\cite{li2022bevsurvey}.

\begin{figure}[ht]
    \centering
    \includegraphics[width=0.95\linewidth]{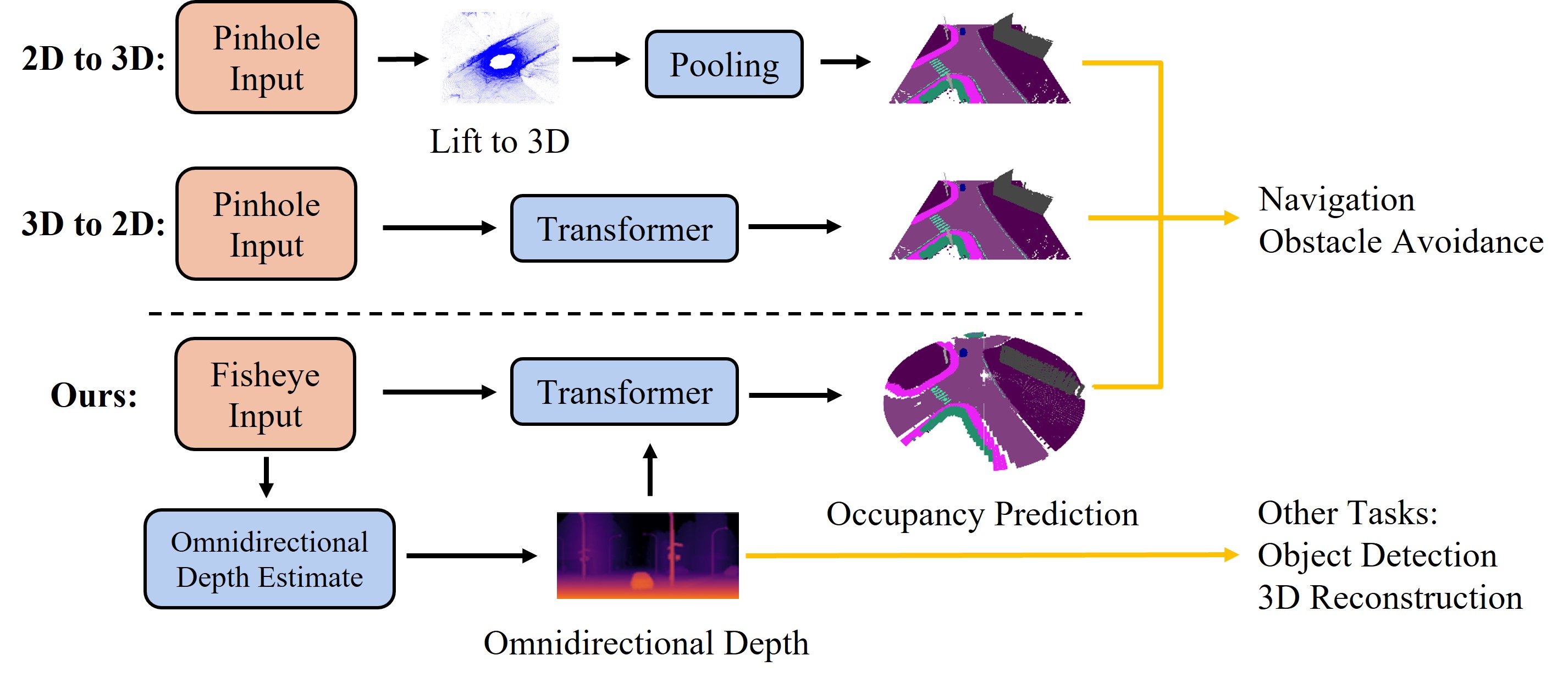}
    \caption{The typical workflow of occupancy prediction \& our Sketch-Coloring framework.}
    \label{fig:workflow}
    \vspace{-1.0em}
    
\end{figure}

The methods for view transformation mainly fall into two categories, as shown in Fig.\ref{fig:workflow}. The first involves using explicit depth estimation to elevate 2D features from images into a 3D voxel space~\cite{philion2020lift,huang2021bevdet,huang2022bevdet4d,li2022bevdepth}. However, this method is difficult to reconstruct occluded regions and thus producing suboptimal results. The second category is based on transformer architecture. Spatial-Cross-Attention (SCA) is used to extract 3D features from the images~\cite{li2023voxformer,sima2023occnet, huang2023tpv}. However, transformer-based methods rely solely on attention mechanism for perspective transformation without any geometric prior. This makes it challenging to distinguish between image regions and the spatial relationship of real-world objects~\cite{li2023voxformer}, leading to a heavy dependence on large datasets. To address these limitations, using depth estimation result to constrain the transformer is necessary.

Moreover, traditional occupancy (Occ) schemes typically use multiple pinhole cameras mounted around the vehicle. Due to the limited field of view (FOV) of pinhole cameras, the overlapping regions between adjacent views are usually very small. In this scenario, the process of extracting depth information from images tends to rely on monocular depth estimation, which is an ill-posed problem. The accuracy of monocular depth estimation is significantly inferior to that of omnidirectional depth estimation with more overlapping regions~\cite{won2020omnimvs}, severely impacting the effectiveness of scene completion. 

\begin{figure}[ht]
    \centering
    \subfigure[$v_a$ and $v_b$ are equidistant from the ego vehicle, with $v_a$ within the voxel grid and $v_b$ outside.]{
    \begin{minipage}[b]{0.17\textwidth}
    \includegraphics[width=1\textwidth]{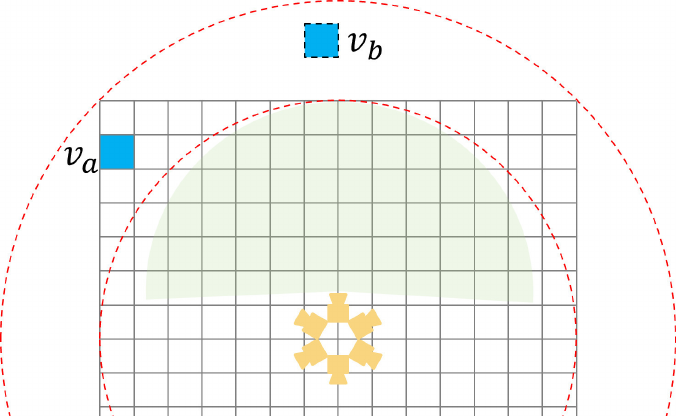}
    \label{fig:cyl}
    \end{minipage}
    }
    \hspace{0.1in}
    \subfigure[An explanation of the cylindrical voxel volume.]{
    \begin{minipage}[b]{0.23\textwidth}
    
    \includegraphics[width=1\textwidth]{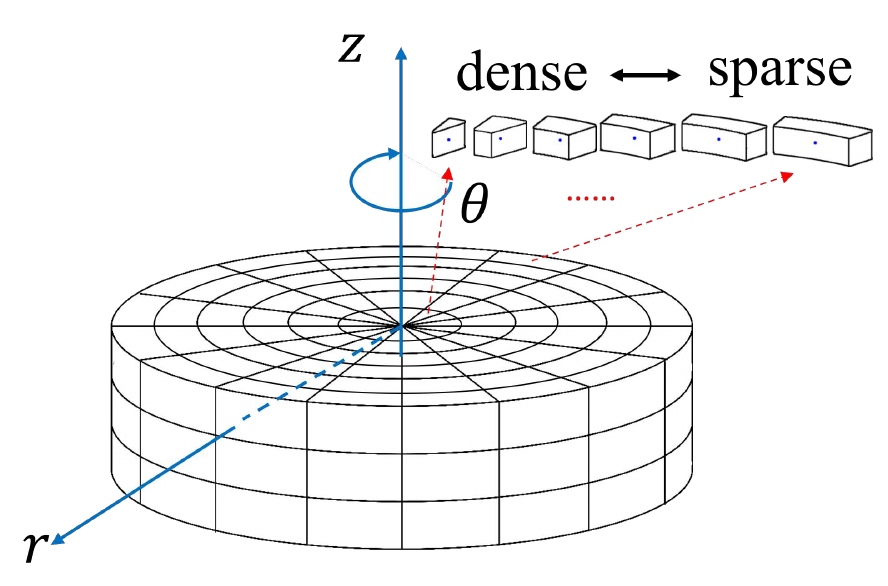}
    \label{fig:coor}
    \end{minipage}
    }
    \caption{Two instances to illustrate the characteristic of Cartesian and Cylinder coordinate system. }
    \label{fig:pro}

\end{figure}

Currently, most occupancy prediction methods are based on the Cartesian coordinate system, where space is uniformly divided to obtain a cuboid voxel grid. Each voxel contains both occupancy and semantic information. However, we have identified several issues with this voxel-based approach:
\begin{itemize}
\item \textbf{A mismatch between voxel resolution and prediction accuracy.}  All voxels are the same size, but in practice, voxels closer to the ego vehicle have higher semantic confidence than those farther away. In this context, the prediction accuracy for nearby regions is constrained by voxel resolution, while prediction error for distant regions often exceeds the size of the voxels. This leads to results that do not accurately represent the real-world scene.

\item \textbf{A mismatch between the perception field and surround-view cameras.}  For instance, consider two voxels, $v_a$ and $v_b$, which are equidistant from the ego vehicle. But $v_a$ is within the voxel grid and $v_b$ is outside it, as illustrated in Fig.\ref{fig:cyl}. For the input image, $v_a$ and $v_b$ have the same weight (they share the same depth). But only the features of $v_a$ will contribute to the loss function, while $v_b$ will not, which hinders network convergence.
\item \textbf{Unreasonable voxel density allocation.} For a moving vehicle, nearby objects are more crucial to monitor than distant ones since they pose a greater immediate threat. Downstream tasks like obstacle avoidance and navigation should focus on nearby objects. Thus, a 3D spatial representation for autonomous driving should be denser in nearby regions to align with the practical needs of autonomous driving scenarios.
\end{itemize}

\begin{figure}[ht]
    \centering
    \includegraphics[width=0.8\linewidth]{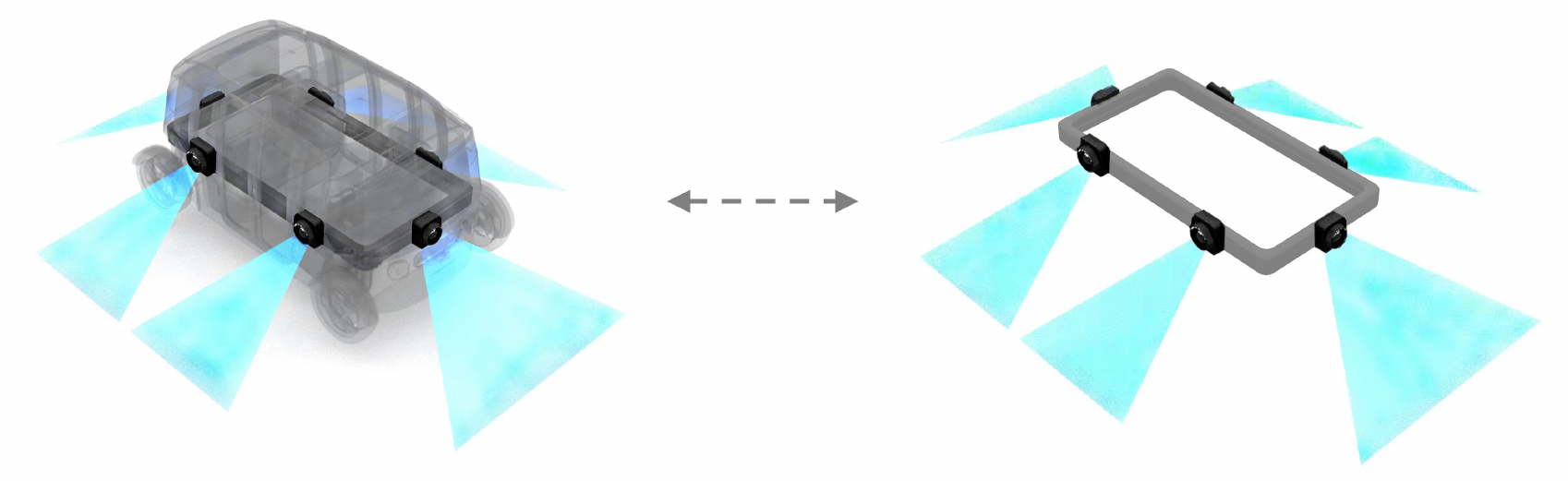}
    \caption{The distribution of cameras on our unmanned vehicle.}
    \label{fig:car}
    \vspace{-1.5em}
\end{figure}
    

Motivated by the aforementioned insights, we propose a Sketch-Coloring framework for occupancy prediction tasks in a polar coordinate system using multiple fisheye cameras. We design a cylindrical voxel grid, as shown in Fig.\ref{fig:coor}, which is more suitable and natural for occupancy prediction tasks compared to Cartesian coordinate system. In fact, some lidar-based 3D object detection methods have already leveraged the polar coordinate~\cite{jiang2022polarformer,PolarDETR,Zhang_2020_CVPR,zhu2020cylinder3d,DBLP:journals/corr/abs-2005-09927}. Our method leverages depth estimation to predict approximate geometric information, leading to higher accuracy in occupancy prediction. By leveraging fisheye cameras' overlapping regions, this approach boosts depth estimation accuracy and enhances scene completion performance. We used an unmanned vehicle equipped with six fisheye cameras to validate our method in a virtual environment. The specific camera distribution is shown in Fig.\ref{fig:car}. It should be noted that the distribution of our cameras can be flexibly adjusted according to requirements.

Our contributions are as follows.
\begin{enumerate}
    \item \textbf{A Sketch-Coloring network structure based on depth estimation prior.} Leveraging larger overlapping regions between fisheye cameras can greatly improve depth estimation accuracy and provide precise structural information for the Sketch-Coloring network. The experimental results show a significant improvement with our method.
    \item \textbf{A cylindrical voxel representation inspired by the polar coordinate to boost prediction performance.} This representation adjusts the density of the voxel on the basis of distance. It better aligns with surround-view cameras' distribution characteristics and enhances prediction performance. 
    \item \textbf{A virtual scene dataset with six fisheye cameras.} To address the lack of fisheye camera data in autonomous driving tasks, we developed a virtual scene dataset using six fisheye cameras.
\end{enumerate}

\section{Related Work}

\subsection{Occupancy Datasets}
\textbf{Data Modality.} Most current Occ datasets such as SemanticKITTI~\cite{behley2019semantickitti}, Occ3D~\cite{tian2023occ3d}and OpenOccupancy~\cite{wang2023openoccupancy} have RGB inputs in the form of pinhole, and the semantic truth values are usually composed of cuboid semantic voxels. There are few occupancy prediction datasets that include fisheye images with large FOV, and our dataset provides a novel surround-view fisheye RGB input modality.

\textbf{Acquisition Method.} At present, there are two main ways to collect SSC datasets for autonomous driving scenarios: 1) Collecting data in real driving scenarios requires the construction of a complete hardware system, and the post-processing costs high, just as SemanticKITTI~\cite{behley2019semantickitti}, Occ3D~\cite{tian2023occ3d}; 2) Collecting data in simulation scenarios has the advantages of free configuration of sensors and easier post-processing, just as Paris-CARLA-3D~\cite{Deschaud2021ParisCARLA3DAR}, CarlaSC~\cite{carlasc}.

\subsection{3D Scene Representation}

\textbf{BEV.} The past several years have witnessed the prosperous development of BEV representation in tasks such as 3D object detection~\cite{li2022bevformer,Yang2022BEVFormerVA,philion2020lift,li2022bevdepth,liu2022petr,liu2022petrv2}, BEV semantic segmentation~\cite{peng2022bevsegformerbirdseyeview,roddick2020predictingsemanticmaprepresentations, zhou2022cross}. Despite efficiency, the geometry lossy projection results in the relatively coarse representation of the 3D scene, hindering its generalization to the fine-grained semantic occupancy prediction task.

\textbf{Cartesian Coordinate.} Representing the surrounding environment with spatial latent information is a trend for autonomous driving perception algorithms. A direct approach is dividing the scene into a cuboid voxel grid of size $(H,W,Z)$ in Cartesian coordinate, as seen in methods~\cite{cao2022monoscene,wei2023surroundocc,ma2023cotr,boeder2024occflownetselfsupervisedoccupancyestimation,ma2024cvpr,wang2023panoocc,sima2023occnet,jiang2023symphonies,zhang2023occformer}. Similarly, in LiDAR point cloud segmentation tasks, the point cloud is often divided into a cuboid voxel grid for subsequent feature encoding and decoding~\cite{wang2023openoccupancy,roldao2020lmscnet,yan2021js3c,song2016ssc}.


\textbf{Polar Coordinate.} In LiDAR-based perception methods, polar or quasi-polar coordinate systems have been explored for 3D object detection. For example, Cylinder3D~\cite{zhu2020cylinder3d} proposes a polar coordinate-based segmentation method for LiDAR point clouds. PointOcc~\cite{pointocc} introduces a cylindrical coordinate-based tri-view representation of point clouds. In vision-based perception methods,
PolarDETR~\cite{PolarDETR} presents polar parameterization for 3D detection and reformulates position parameterization.
\subsection{Vision-Based Occupancy Prediction}
Occupancy prediction networks focus on reconstructing voxel-level 3D scene and predicting voxel-level semantic information from input images. MonoScene~\cite{cao2022monoscene} achieves scene occupancy completion through a 2D and a 3D UNet connected by a sight projection module. OccNet~\cite{sima2023occnet} applies universal occupancy features to various downstream tasks and introduces the OpenOcc benchmark. 
To address the computational burden of handling a large number of voxel queries, TPVFormer~\cite{huang2023tpv} proposes using tri-perspective view representation to supplement vertical structural information, though this inevitably leads to information loss. 
VoxFormer~\cite{li2023voxformer} initializes sparse queries based on stereo depth prediction. Several methods emerged in the CVPR 2023 occupancy challenge~\cite{pan2023renderocc,ding2023multiscaleocc4thplace,li2023fbocc,pan2023uniocc}, but none explored voxel representation in polar coordinate or leveraged panoramic depth estimation to introduce geometric prior.
\begin{figure*}[t]
    \centering
    \vspace{1.0em}
    \includegraphics[width=0.90\linewidth]{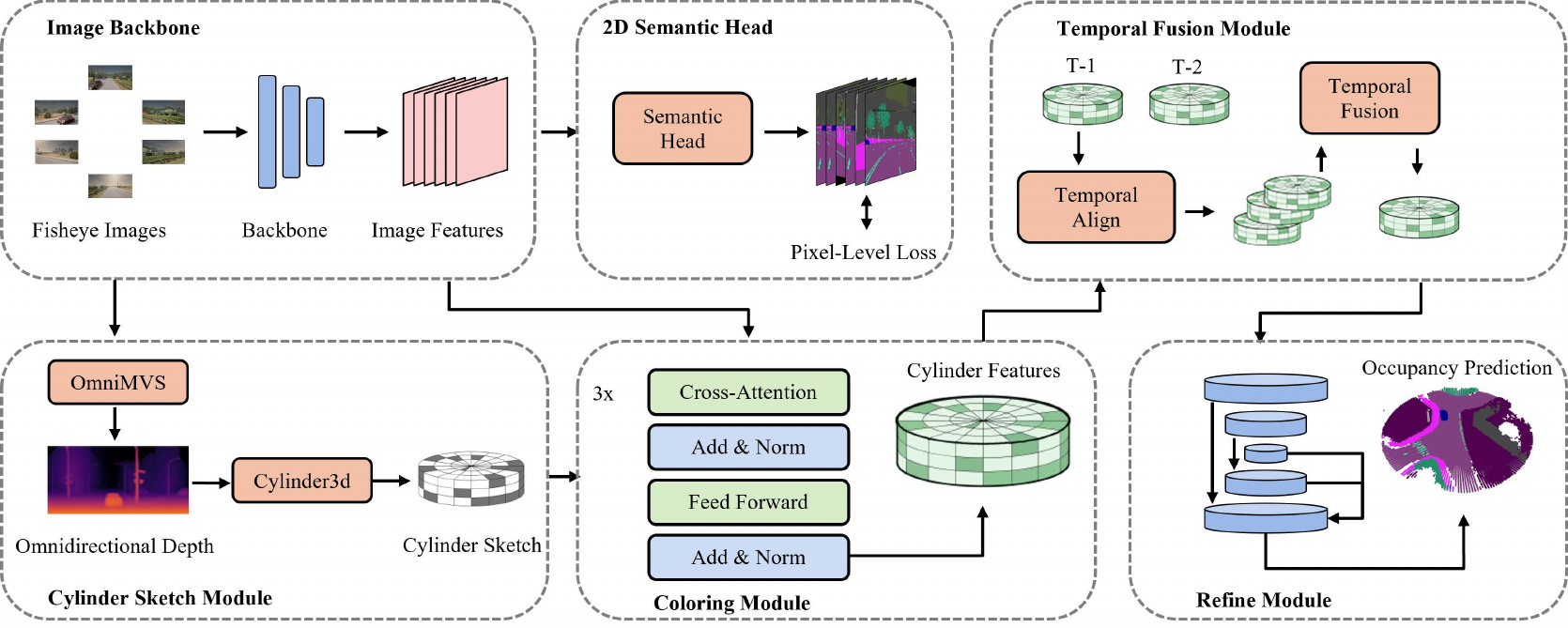}
    \vspace{-1.0em}
    
    \caption{Overall architecture}
    \label{fig:overall}
    \vspace{-2.0em}
\end{figure*}
\section{Virtual Scene Occ Dateset with Fisheye Camera Input}

To collect an occupancy dataset from fisheye cameras, we build a simulation data collection system based on the open-source autonomous driving simulator Carla Simulator~\cite{Dosov17carla}. This system collects multi-view RGB images and semantic label ground truth. The distinctions of this dataset from other occupancy datasets are shown in Table \ref{tab:dataset}.
\begin{table}[ht]

\setlength{\tabcolsep}{3pt}
\caption{Compare against classic Occ datasets}
\vspace{-1.0em}

\label{tab:dataset}
\begin{center}
\scriptsize
\begin{tabular}{ccccccccc}
\toprule
method&\makecell{pinhole\\images}&\makecell{fisheye\\images}&\makecell{cuboid\\voxels}&\makecell{polar\\voxels}&\makecell{2D semantic\\ \&depth}&\makecell{frames}\\
\midrule
SemanticKITTI~\cite{behley2019semantickitti} &\checkmark   &          &         \checkmark &	&       & 23,201    \\
Occ3D-nuScenes~\cite{tian2023occ3d}        &\checkmark	&          &        \checkmark  &	&		   & 40,000   \\
CarlaSC~\cite{carlasc} &\checkmark	&&    \checkmark   	  & \checkmark& 	      & 43,200\\
Ours&\checkmark	&\checkmark&\checkmark&	\checkmark&	\checkmark& 57,600\\

\bottomrule
\end{tabular}
\end{center}
\vspace{-2.0em}
\end{table}

Inspired by CarlaSC~\cite{carlasc}, our dataset collection utilizes Carla Simulator to run a virtual driving scene in UE4 and collect data by various sensors.

It is worth mentioning that we collect cubemap images and post-process them into fisheye images. We place 20 semantic LiDARs around the vehicle and aggregate the semantic point clouds collected by all semantic LiDARs in each frame to obtain a complete semantic point cloud. We then sample this point cloud with voxel grid (both cuboid and cylindrical), then each voxel’s semantic label is determined by voting among the semantic points within it.

From the perspective of scene collection, We have collected 32 scenes under various weather conditions across 8 different maps, capturing 1800 frames per scene at a frame rate of 10 FPS. In total, the dataset comprises 57,600 frames of data. Our dataset encompasses a variety of scene conditions, including sunny, rainy, and overcast skies, and is twice the size of Semantic KITTI~\cite{behley2019semantickitti}. If our paper is accepted, we will open-source our dataset to the community.

\section{Methodology}

 \begin{figure}[ht]
    \centering
    \subfigure[Dilation window varying with distance.]{
    \begin{minipage}[b]{0.15\textwidth}
    \includegraphics[width=\textwidth]{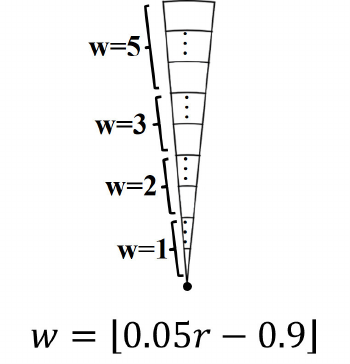}
    \label{fig:dila_a}
    \vspace{-1em}
    \end{minipage}
    }
    \hspace{0.1in}
    \subfigure[An example to illustrate the dilation process when the dilation window is 2.]{
    \begin{minipage}[b]{0.25\textwidth}
    \includegraphics[width=1\textwidth]{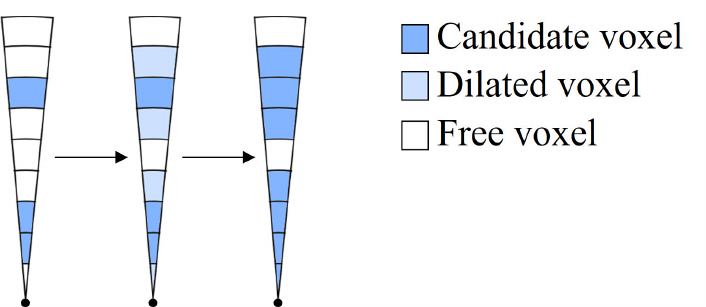}
    \label{fig:dila_b}
    \end{minipage}
    }
    \caption{The dilation process after the Cylinder Sketch Module}
    \label{fig:dila}
    \vspace{-1.5em}
    
\end{figure}
\subsection{Overall Architecture}

As shown in Fig.\ref{fig:overall}, our method consists of six main components: Image Backbone, Cylinder Sketch Module, 2D Semantic Head, Coloring Module, Temporal Fusion, and Refine Module. The Cylinder Sketch Module predicts a cylindrical, class-agnostic occupancy grid using omnidirectional depth estimation, akin to sketching. The Coloring Module uses a transformer to map image features onto the cylindrical voxel grid, akin to coloring. The 2D Semantic Head performs semantic segmentation to enhance the backbone's semantic learning. Temporal Fusion combines temporal information from multi-frame inputs, and the Refine Module further refines voxel features. Each component is detailed in the following sections.

\subsection{Cylinder Sketch Module}

\textbf{Omnidirectional Depth Estimation.} For omnidirectional depth estimation, we employ RtHexa-OmniMVS~\cite{li2024realtime}. RtHexa-OmniMVS projects the feature maps from fisheye images onto N spherical surfaces with varying radius, then the matching cost is calculated to estimate depth. The obtained spherical depth is transformed to ERP (Equirectangular Projection) depth image.
Then we utilize the camera's intrinsic and extrinsic parameters to convert the ERP depth map into a pseudo-point cloud.

\textbf{Cylinder Sketch.} Methods based on transformers often struggle with convergence due to a lack of geometric constraints. To address this, we design the Cylinder Sketch Module to predict a sparse set of candidate voxels, which participate in the query process, to save memory. This approach is like sketching an outline. We divide the pseudo point cloud from depth estimation into cylindrical voxels and use the Cylinder3D~\cite{zhu2020cylinder3d} framework for candidate prediction. Additionally, we dilate along the cylindrical sketch's radius to reduce depth estimation blurriness in distant regions, considering voxels near an occupied one within the dilation window as occupied.This dilation method is like the dilation operation in image processing.

Fig.\ref{fig:dila} illustrates the dilation process with a dilation window of 2. We set different dilation window for different distance ranges, with larger dilation window for farther regions, as shown in Fig.\ref{fig:dila_a}.

\subsection{Coloring Module} 

We design a Coloring Module to infuse the sketch with semantic features, akin to coloring a sketch. This module primarily consists of a cross attention structure, where we use the Cylinder Sketch as candidate query voxels. Specifically, we locate the corresponding 2D reference points of the candidate voxels by camera's intrinsic and extrinsic parameters. These points serve as reference points for Deformable Cross Attention (as in DETR~\cite{detr3d}). The formula for Coloring Module(CM) is as follows:
\begin{equation}
\label{eq:cm}
    CM(q_c,F)=\frac{1}{H}\sum_{i\in{H}}DA(q_c,P(p,T_i),F_i)
\end{equation}
Here, $\text{DA}$ is the Deformable Attention module, $q_c$ is the query corresponding to the candidate voxel, $F$ is the image feature, and $P(p, T_i)$ represents the normalized pixel coordinate in the image obtained based on the voxel position and camera parameters. $H$ includes all the feature maps that the candidate voxels hit when projected onto the image features. This process effectively lifts 2D features into 3D features, improving computational efficiency and feature representation accuracy. 

Due to their wide field of view, fisheye cameras introduce significant radial and tangential distortions. To address these distortions, we apply the projection method specifically designed for fisheye image inputs and introduce Deformable Convolutional Networks(DCN) in image backbone.

In summary, by performing feature transformation only on the candidate region, we significantly reduce the computational load while ensuring accuracy. Additionally, the projection method tailored for fisheye camera distortions ensures the accuracy of feature mapping in a wide field of view.

\subsection{2D Semantic Head}

 \begin{figure}[ht]
    \centering
    \subfigure[The projection of cuboid voxels.]{
    \begin{minipage}[b]{0.2\textwidth}
    \includegraphics[width=1\textwidth]{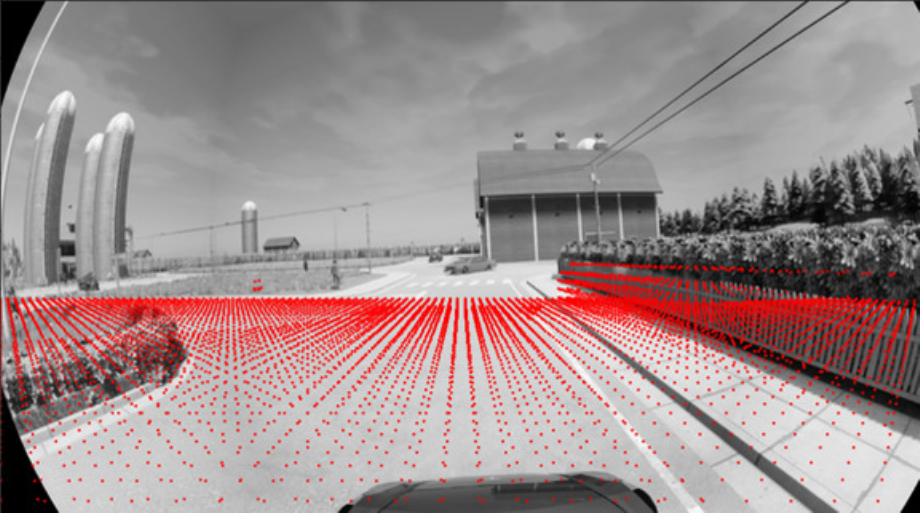}
    \label{fig:map_a}
    \end{minipage}
    }
    \hspace{0.1in}
    \subfigure[The projection of cylindrical voxels.]{
    \begin{minipage}[b]{0.2\textwidth}
    \includegraphics[width=1\textwidth]{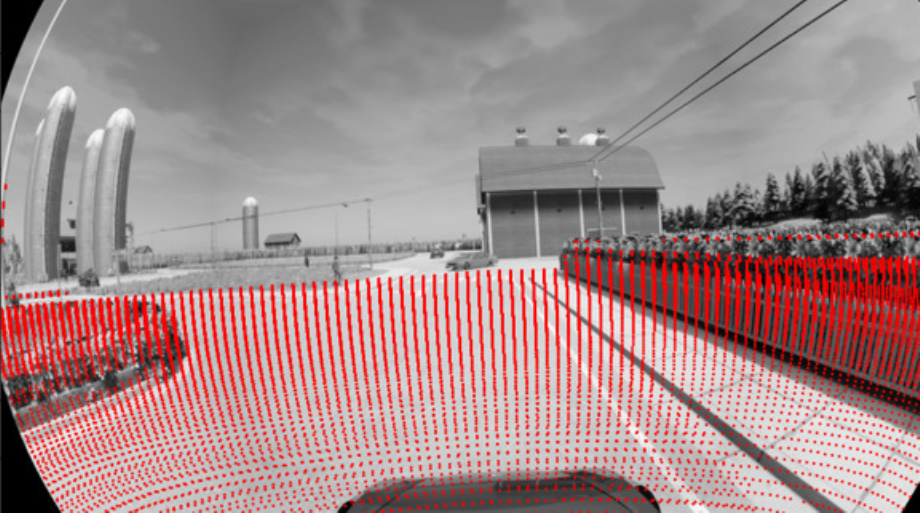}
    \label{fig:map_b}
    \end{minipage}
    }
    \caption{The projection of candidate voxels in the input image reflects the sparsity and imbalance of projection points. The red dots indicate the projection locations of the voxels. }
    \label{fig:map}
    \vspace{-1.5em}
    
\end{figure}
In a typical occupancy task, the supervision is applied to the output voxel grid, where each voxel aggregates 3D information from the corresponding 2D pixels of multiple view images. The loss is calculated between these voxels and the ground truth, which we refer to as voxel-level supervision. We identify two issues that arise from relying solely on voxel-level supervision:
\begin{itemize}

\item \textbf{Indirect Supervision on Images:} The loss directly affects the voxel features. After 3D-to-2D projection and image feature sampling, the loss becomes indirect.

\item \textbf{Sparse Supervision on Image Features:} Since our queries originate from occupancy prediction and are inherently sparse, this sparsity persists through the 3D-to-2D projection. As shown in Fig.\ref{fig:map}, only the sparse pixels around the 2D reference points receive a supervision signal, resulting in a decline in prediction accuracy.

\end{itemize}

Due to these reasons, the sparse voxel-level supervision on image features fails to adequately guide the backbone encoding these features. To address this, we design a semantic segmentation head applied to the image feature backbone. The segmentation head predicts per-pixel semantic labels to calculate the loss, providing richer signals for supervision.

\begin{figure*}[ht]
\vspace{1.0em}
    \centering
    \includegraphics[width=0.9\linewidth]{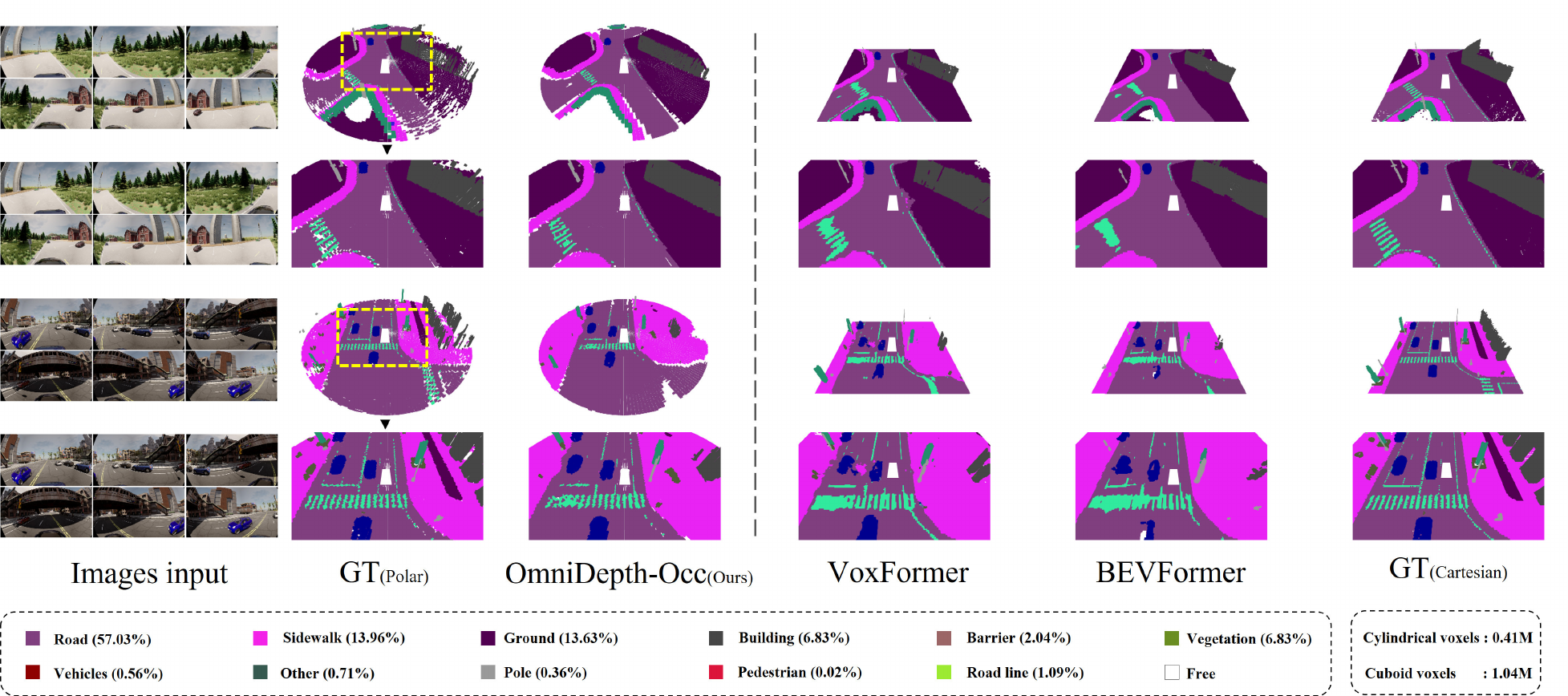}
    \caption{Visualization results}
    \label{fig:compare}
    \vspace{-1em}
    
\end{figure*}
\subsection{Temporal Fusion}
As static images without temporal cues struggle to infer the speed of moving objects or detect highly occluded objects. Therefore, we design a temporal fusion module for 3D voxel queries to further enhance voxel representations.

The temporal fusion module integrates historical voxel query information into the current voxel grid $V_\text{curr}$. As shown in Fig.\ref{fig:overall}, before performing temporal fusion, voxel alignment is crucial for accurately perceiving the environment, which is formulated as follows:
\begin{equation}
\label{eq:fusion}
    V_\text{fusion}=\frac{1}{N+1}(V_\text{curr}+\sum_{i=1}^N\text{GS}(V_i,T_i,T_\text{curr}))
\end{equation}
Here, $N$ denotes the number of frames to be fused, $V_\text{curr}$ is the current frame's voxel features extracted through Coloring Module, and $\text{GS}(V_i,T_i,T_\text{curr})$ represents the historical voxel features obtained using the transformation matrix between the current and historical frames followed by grid sampling.
\subsection{Refine Module}

Following the Coloring Module, we design a refine module, which is a UNet-like encoder-decoder network. After the initial voxel features are extracted by the Coloring Module, the refine module further refines these voxel features. At this stage, the feature extraction is not limited to the candidate voxels but applies to the entire voxel grid. This module plays a vital role in reconstructing detail and leverages the spatial consistency of voxel semantics while reducing noise.
\subsection{Training Loss}

In this paper, the sum of four loss functions is employed as the overall loss function for training purposes,as shown in Eq.\ref{eq:loss}.
\begin{equation}
\label{eq:loss}
    Loss=L_{CE}+L_{scal}+L_{dice}+L_{2Dsem}
\end{equation}
Where $L_{CE}$ denotes the weighted cross-entropy function, the weight for each class can be derived from Eq.\ref{eq:weight}. $L_{scal}$ is inspired by~\cite{cao2022monoscene}, used to mitigate the influence of class imbalance during training.
\begin{equation}
\label{eq:weight}
    \omega_c=\frac{1}{log(f_c+c)}
\end{equation}
As shown in Eq.\ref{eq:dice}, $L_{dice}$ serving as a loss function frequently utilized in image segmentation tasks. 
\begin{equation}
\label{eq:dice}
    L_{dice}=1-\frac{2TP}{2TP+FP+FN}
\end{equation}
The merit of Dice Loss lies in its direct optimization of the F1 score, thereby adeptly tackling extreme cases of class imbalance.
\begin{table*}[htp]
\vspace{0.6em}
\setlength{\tabcolsep}{3.9pt}
\caption{Quantitative comparison against camera-based methods}
\vspace{-2.0em}
\label{tab:compare}

\begin{center}
\begin{tabular*}{1\linewidth}{cccccccccccccc}

\toprule
method&Voxels&RayIoU&Road&Sidewalk&Ground&Building&Wall&Vegetation&Vehicles&Other&Pole&Pedestrian&Roadline\\
\midrule
\textbf{BEVDet~\cite{huang2021bevdet}}   &1.04M&0.310&0.714&0.608&0.505&0.141&0.203&0.095&0.367&0.120&0.113&0.007&0.235\\
\textbf{BEVFormer~\cite{li2022bevformer}}&1.04M&0.301&0.723&0.685&0.518&0.140&0.259&0.137&0.354&0.125&0.142&0.000&0.225\\
\textbf{VoxFormer~\cite{li2023voxformer}}&1.04M&0.388&0.805&0.718&0.258&0.213&0.339&0.226&0.447&0.287&0.326&\textbf{0.338}&0.308\\
\textbf{OmniDepth-Occ(ours)}&\textbf{0.41M}&\textbf{0.470}&\textbf{0.813}&\textbf{0.793}&\textbf{0.606}&\textbf{0.237}&\textbf{0.417}&\textbf{0.373}&\textbf{0.556}&\textbf{0.306}&\textbf{0.368}&0.317&\textbf{0.380}\\
\bottomrule
\end{tabular*}
\end{center}
\vspace{-2.0em}
\end{table*}
\section{Experiment}
\subsection{Experimental Setup}

\textbf{Our Fisheye Dataset.} Based on the proposed six-fisheye camera dataset, we conduct subsequent experiments. The semantic ground truth is in the form of cylindrical coordinate voxels with a shape of  $(128\times200\times16)$, representing $(r,\theta,z)$ in the cylindrical coordinate. We set the perception radius within $(0m, 25.6m)$ and the height range within $(-2.8m, 3.6m)$. In terms of data augmentation, we primarily perform random cropping on the input RGB images to enhance robustness against occlusion.

\textbf{Evaluation Metrics.} Our evaluation metric follows the $RayIoU$ proposed in ~\cite{liu2024rayiou}. It is calculated by projecting query rays into the predicted 3D occupancy volume and ground truth separately. For each query ray, we compute the distance it travels before intersecting any surface and retrieve the corresponding class label to determine whether the prediction matches the ground truth. $RayIoU$ provides a better evaluation in the presence of object occlusions and does not require annotation of the effective region of the camera. The final $RayIoU$ calculation is given by Equation \ref{eq:rayiou}, where $C$ is the total number of classes:
\begin{equation}
\label{eq:rayiou}
    RayIoU=\frac{1}{C}\displaystyle\sum_{c=1}^C\frac{TP_c}{TP_c+FP_c+FN_c}
\end{equation}
Here, $TP$ refers to the cases where the predicted label matches the true label, and the difference in distance between them is within a specified threshold.

\begin{table}[ht]
\setlength{\tabcolsep}{3.9pt}

\caption{result in different distance}
\vspace{-2.0em}
\label{tab:range}
\begin{center}
\begin{tabular}{cccc}
\toprule
method&$\text{RayIoU}_\textbf{0-8.5m}$&$\text{RayIoU}_\textbf{8.5-17m}$&$\text{RayIoU}_\textbf{{17-25.6m}}$\\
\midrule
\textbf{BEVDet}&  0.435   &  0.321   &  0.150\\
\textbf{BEVFormer}&0.421&0.319&0.177\\
\textbf{VoxFormer}&0.454&0.413&0.285\\
\textbf{OmniDepth-Occ(ours)}&\textbf{0.496}&\textbf{0.446}&\textbf{0.294}\\
\bottomrule
\end{tabular}
\end{center}
\vspace{-1.0em}
\end{table}

\begin{table}[htp]
\caption{Ablation Study for Semantic Head \& Temporal Fusion}
\vspace{-1.5em}
\label{tab:ablation}
\begin{center}
\begin{tabular}{cc}
\toprule
method&RayIoU\\
\midrule
ours&\textbf{0.470}\\
w/o 2D semantic head&0.443\\
w/o temporal fusion&0.458\\
\bottomrule
\end{tabular}
\end{center}
\vspace{-2.0em}
\end{table}

\textbf{Implementation Details.} For panoramic depth estimation, we employ RtHexa-OmniMVS, which generate panoramic depth maps. Our Cylinder Sketch Module is inspired by Cylinder3D ~\cite{zhu2020cylinder3d}, a highly efficient LiDAR semantic segmentation network. We train Cylinder3D for $30$ epochs on 4 NVIDIA 3090 GPUs. We employ a ResNet-50 backbone with DCN as image backbone. The extracted features are fed into an FPN to aggregate multi-scale image features. The Coloring Module utilizes a Cross-Attention structure with a feature dimension of $d=128$ for the query. For the temporal fusion module, we choose the three frames preceding the current frame. The voxel features after temporal fusion are fed into the refine module. Subsequently, a linear layer compresses the output feature dimension to $12$, corresponding to the total number of classes. We train the model for $40$ epochs on 4 NVIDIA 3090 GPUs, with a learning rate of $2 \times 10^{-4}$.

\subsection{Occupancy Prediction Result}

In Table \ref{tab:compare}, we present the quantitative results on our validation set. All methods use only camera input and are trained for 40 epochs. Since VoxFormer~\cite{li2023voxformer} typically uses monocular or stereo depth along with single-view images, we extend it to a panoramic input and retrain it on our pinhole data. These methods all employ the same backbone ResNet-50.

Our method shows a 21\% improvement compared to VoxFormer, a 56\% improvement compared to BEVFormer, and a 51.6\% improvement compared to BEVDet. Furthermore, we conduct a quantitative evaluation over different distance ranges, as shown in Table \ref{tab:range}. In comparison to VoxFormer, our model has realized more significant advancements in near-field performance, a benefit we ascribe to the augmented resolution in proximate ranges coupled with more precise depth estimation. As illustrated in Figure \ref{fig:compare}, compared to other classic methods, our method achieves finer scene reconstruction at all ranges and more accurately reconstructs dynamic objects, such as moving vehicles, with minimal trailing artifact (0.447-0.556). The overall performance is significantly enhanced.

\begin{table}[htp]
\caption{Ablation Study for Refine Module}
\vspace{-1.5em}
\label{tab:refine}
\begin{center}
\begin{tabular}{cc}
\toprule
method&RayIoU\\
\midrule
our refine module&\textbf{0.470}\\
self-attention&0.456\\
w/o refine module&0.391\\ 
\bottomrule
\end{tabular}
\end{center}
\vspace{-1.0em}
\end{table}

\subsection{Ablation Study}

Ablation studies on temporal fusion and joint semantic supervision are presented in Table \ref{tab:ablation}. Compared to the case of using only the current frame input, our strategy of fusing three historical frames improves the mIoU by 2.6\%(0.458-0.470). Compared to the case without 2D semantic supervision, our structure improves the mIoU by 6\% (0.443-0.470).

Refine module ablation studies are presented in Table \ref{tab:refine}. Our structure shows an 3\% improvement (0.456-0.470) over the self-attention structure from VoxFormer, and improves the mIoU by 20\% (0.391-0.470) compared to the method without refine module.
\begin{table}[t]
\caption{Ablation Study for Cylinder Sketch Module}
\vspace{-1.0em}
\label{tab:sketch}
\begin{center}
\begin{tabular}{cc}
\toprule
method&RayIoU\\
\midrule
omni depth&\textbf{0.470}\\
omni depth w/o dilation&0.465\\
mono depth&0.263\\
dense voxels&0.434\\
random voxels&0.401\\

\bottomrule
\end{tabular}
\end{center}
\vspace{-2.0em}
\end{table}
Cylinder sketch module ablation studies are presented in Table \ref{tab:sketch}.
Results show that our cylinder sketch module, based on panoramic depth estimation, achieves a 78\% improvement (0.263-0.470) compared to inputs using monocular depth estimation. It also shows an 8.2\% improvement (0.434-0.470) over fully dense voxels and a 17.2\% improvement (0.401-0.470) compared to randomly distributed voxels (with a candidate voxel ratio of 25\%). Additionally, our dilation operation contributes to a slight enhancement in the results (0.465-0.470).

We conducted ablation experiments on the cylindrical voxel representation, with the results presented in Table \ref{tab:repre}. During the experiments, the refine module and temporal fusion module were removed, and only the effect of replacing square voxels with cylindrical voxels was observed. The experimental results indicate that the predicted RayIoU increased from 0.388 to 0.454(17.0\%). 
\begin{table}[t]
\caption{Ablation Study for cylindrical voxel representation}
\vspace{-1.0em}
\label{tab:repre}
\begin{center}
\begin{tabular}{cc}
\toprule
method&RayIoU\\
\midrule
cylindrical voxel representation&\textbf{0.454}\\
cuboid voxel representation&0.388\\

\bottomrule
\end{tabular}
\end{center}
\vspace{-2.0em}
\end{table}

\section{Conclusion}

In summary, we present OmniDepth-Occ, a cylindrical voxel-based Sketch-Coloring framework. It uses omnidirectional depth estimation to constrain transformer. Additionally, we propose a polar coordinate-based cylindrical voxel representation and a virtual scene dataset with six fisheye cameras for our network. Experimental results demonstrate that our OmniDepth-Occ achieves a significant accuracy improvement especially in close range, surpassing classical methods like VoxFormer and BEVFormer. Our system not only generates occupancy prediction for navigation and obstacle avoidance, but also provides dense depth information that can offer sufficient structural information for more tasks such as 3D reconstruction and object detection. These will be further explored in our future work.
\clearpage
\clearpage
\addtolength{\textheight}{-12cm}   




\bibliography{reference.bib}
\end{document}